\documentclass{article}

\usepackage{microtype}
\usepackage{graphicx}
\usepackage{subfigure}
\usepackage{booktabs} 
\usepackage{bera}
\usepackage{listings}
\usepackage{xcolor}
\usepackage{multicol}
\usepackage{filecontents}
\usepackage{lipsum}
\usepackage{amsmath}

\colorlet{punct}{red!60!black}
\definecolor{background}{HTML}{EEEEEE}
\definecolor{delim}{RGB}{20,105,176}
\colorlet{numb}{magenta!60!black}

\lstdefinelanguage{json}{
    basicstyle=\normalfont\ttfamily,
    showstringspaces=false,
    breaklines=true,
    frame=lines,
    backgroundcolor=\color{background},
    literate=
     *{0}{{{\color{numb}0}}}{1}
      {1}{{{\color{numb}1}}}{1}
      {2}{{{\color{numb}2}}}{1}
      {3}{{{\color{numb}3}}}{1}
      {4}{{{\color{numb}4}}}{1}
      {5}{{{\color{numb}5}}}{1}
      {6}{{{\color{numb}6}}}{1}
      {7}{{{\color{numb}7}}}{1}
      {8}{{{\color{numb}8}}}{1}
      {9}{{{\color{numb}9}}}{1}
      {:}{{{\color{punct}{:}}}}{1}
      {,}{{{\color{punct}{,}}}}{1}
      {\{}{{{\color{delim}{\{}}}}{1}
      {\}}{{{\color{delim}{\}}}}}{1}
      {[}{{{\color{delim}{[}}}}{1}
      {]}{{{\color{delim}{]}}}}{1},
}

\usepackage{hyperref}

\usepackage[accepted]{packages/sysml2019}

\sysmltitlerunning{Nonlinear Conjugate Gradients for Scaling Synchronous Distributed DNN Training}

\begin{document}

\twocolumn[
\sysmltitle{Nonlinear Conjugate Gradients for Scaling Synchronous Distributed DNN Training}



\sysmlsetsymbol{equal}{*}

\begin{sysmlauthorlist}
\sysmlauthor{Saurabh Adya}{appl}
\sysmlauthor{Vinay Palakkode}{appl}
\sysmlauthor{Oncel Tuzel}{appl}
\end{sysmlauthorlist}

\sysmlaffiliation{appl}{Apple Inc., Cupertino, California, USA}

\sysmlcorrespondingauthor{Saurabh Adya}{sadya@apple.com}

\sysmlkeywords{Distributed Training, Optimization, Deep Learning, Distributed Systems}

\vskip 0.3in

\begin{abstract}\label{abstract}
Nonlinear conjugate gradient (NLCG) based optimizers have shown superior loss convergence properties compared
to gradient descent based optimizers for traditional optimization problems. However, in
Deep Neural Network (DNN) training, the dominant optimization algorithm of choice is still
Stochastic Gradient Descent (SGD) and its variants. In this work, we propose and evaluate the stochastic
preconditioned nonlinear conjugate gradient algorithm for large scale DNN training tasks. We show that a
nonlinear conjugate gradient algorithm improves the convergence speed of DNN training, especially
in the large mini-batch scenario, which is essential for scaling synchronous distributed
DNN training to large number of workers. We show how to efficiently use second order information in the NLCG
pre-conditioner for improving DNN training convergence. For the ImageNet classification task,
at extremely large mini-batch sizes of greater than 65k, NLCG optimizer is able to improve top-1 accuracy by
more than 10 percentage points for standard training of the Resnet-50 model for 90 epochs.
For the CIFAR-100 classification task, at extremely large mini-batch sizes of greater than 16k, NLCG optimizer is able to
improve top-1 accuracy by more than 15 percentage points for standard training of the Resnet-32 model
for 200 epochs.
\end{abstract}

]



\printAffiliationsAndNotice{}  

\section{Introduction}\label{sec:introduction}
As dataset sizes and neural network complexity increases, DNN training times have exploded.
It is common to spend multiple weeks training an industrial scale DNN model
on a single machine with multiple GPUs. Reducing DNN training times using distributed DNN training
\cite{tensorflow:paper} is becoming imperative to get fast and reasonable turn around times for research
experiments and training production DNN models. In addition, faster DNN training methods can reduce the cost
of training.

In this work we study the limits of synchronous DNN training on popular large-scale DNN training tasks:
ImageNet classification \cite{ILSVRC15} and CIFAR-100 classification.
Several system level challenges need to be resolved in order to implement an efficient large-scale synchronous
distributed training algorithm. Once the throughput and latency issues have been optimized, the fundamental limitation to
scaling a synchronous DNN algorithm is making effective use of an extremely large mini-batch.
There has been recent work extending SGD based algorithms to maintain the model accuracy as
the mini-batch size increases \cite{akiba2017extremely}, \cite{goyal2017accurate}, \cite{lars:paper}, \cite{progressivebatch:paper},
\cite{fastai} and they have shown promise for training Resnet-50 on ImageNet.

In this work, we take an orthogonal approach and explore an algorithm that uses the large mini-batch sizes
more effectively than SGD and its variants.
We study the performance of various optimizers in the standard DNN training setting.
We propose to use the preconditioned nonlinear conjugate gradient method \cite{polakribiere}, \cite{fletcherreaves},
\cite{Nocedal:opt} in the standard DNN training setting.
We demonstrate how to efficiently use second order information in the pre-conditioner of the NLCG optimizer.
To our knowledge, this is the first work that shows NLCG based optimizers can provide better solutions than
SGD based optimizers for large scale DNN training tasks like ImageNet, particularly for
large mini-batch training, which is essential for scaling up synchronous distributed training.

In Section \ref{sec:distributed_training}, we review distributed DNN training methods and their challenges.
In Section \ref{sec:nlcg}, we describe the stochastic preconditioned nonlinear conjugate gradient method
and its application to DNN training. In Section \ref{sec:results} we compare NLCG and
SGD based methods for training the Resnet-50 model for the ImageNet classification task and
training the Resnet-32 model for the CIFAR-100 classification task.

\section{Distributed DNN training}\label{sec:distributed_training}

DNN training can be distributed either by splitting the model (model parallelism) or splitting the
data (data parallelism) \cite{dean2012large}. In model
parallelism, the neural network is split across multiple worker nodes. This is typically employed when the
network is too large to fit on a single worker. More common is data parallelism in which each worker
uses different subsets of the data (mini-batches) to train the same model.
When a \emph{parameter server} is used \cite{dean2012large}, the consensus DNN weights are stored on
the parameter server. Meanwhile each worker keeps a copy of the DNN graph and the latest weights.
At each iteration each worker samples a mini-batch of training data and computes an estimate of the
gradient of the loss with respect to the weights. All workers communicate their estimate to the parameter server which
updates the weights and broadcasts the updated weights to all workers.

\subsection{Data Parallel Distributed DNN training algorithms}
One can use different distributed training algorithms to apply the weight update within the data parallel framework.
In asynchronous training methods like asynchronous SGD \cite{dean2012large},
the workers execute without synchronization. Each worker computes an update to the weights
independently and on different mini-batches. The update is communicated to the parameter server which
updates the weights and communicates these back to the worker.
The weight update happens asynchronously. Since there is no explicit synchronization point the overall
system throughput is high. To further reduce the communication overhead methods like Block
Momentum SGD (BMSGD)\cite{bmsgd:paper} and Elastic Averaging SGD (EASGD)\cite{easgd:paper} have been proposed. These
perform N updates on each worker before communicating an N-step weight update to the
parameter server. These methods increase system throughput by increasing computation to communication ratio,
but introduce additional disparity between the weights and the updates applied.

Synchronous DNN training \cite{dean2012large} works by explicitly synchronizing the workers.
Coordination happens when all the gradient estimates (one per worker) are averaged and applied
to the master model. Synchronous DNN training methods can be implemented using
parameter servers \cite{tensorflow:paper} or techniques from High Performance Computing (HPC) like
ring all-reduce averaging \cite{baidu:hpc}. With synchronous DNN training the effective mini-batch size
increases. A larger mini-batch size has the effect of reducing the variance of the stochastic gradient.
This reduced variance allows larger steps to be taken (increase the learning rate)
\cite{goyal2017accurate}.

It has been shown that for the ImageNet training task, as the number of workers increases,
synchronous distributed methods tend to have better performance than asynchronous methods \cite{revisitsync}.
In our experiments, we also observe that
given a fixed number of epochs, synchronous DNN training has better final accuracy than its asynchronous counterparts.
Hence we focus on optimization methods for scaling of synchronous DNN training as the number of workers increases.
We conduct our research on training the popular Resnet-50 model
\cite{resnet:paper} on the ImageNet image classification task \cite{ILSVRC15}.
We also test our proposed optimization algorithm on training the Resnet-32 model on the
CIFAR-100 image classification task.


\subsection{Scaling Synchronous Distributed DNN training throughput}
The work in \cite{revisitsync} shows that throughput for parameter server based distributed
synchronous training is reduced because of the straggler effect of the slowest worker / communication link
dominating the overall system throughput.
It also suggests that to improve system throughput and mitigate the straggler effect, we should slightly
reduce the number of gradients averaged compared to the number of workers in the cluster.


Synchronous DNN training throughput can be further improved using distributed ring reduce algorithm \cite{horovod:paper},
from High Performance Computing (HPC) field, to  average the gradients from different machines. The Horovod library
\cite{horovod:paper} built on top of Tensorflow provides a good implementation of ring all-reduce using MPI and NCCL
libraries.

\subsection{Large Mini-batch Size DNN Training}
Scaling up synchronous distributed training requires the ability to train with large mini-batches.
In synchronous distributed training, we should keep the per worker
mini-batch size as high as possible to maintain high system throughput (high compute to communication ratio).
With a fixed mini-batch size per worker, the effective mini-batch size per training step becomes larger
as the number of workers increases.
If we keep the number of training epochs constant the number of training steps (weight updates) reduces.
So as we scale the number of workers, to achieve lower training times,
the DNN needs to train in fewer training steps. Until a certain mini-batch size, this can be achieved effectively
by increasing the learning rate proportionally
to the mini-batch size \cite{goyal2017accurate}. Different research groups have observed that at some problem specific
mini-batch size, accuracy of the DNN training starts reducing due to training not having converged to an optimal point
\cite{goyal2017accurate}, \cite{dataparallel}.
Figure \ref{fig:accuracy_vs_batchsize} shows this effect for training Resnet-50 model on ImageNet dataset
where the optimization fails to converge to baseline accuracy in 90 epochs for mini-batch sizes above 32768.
As suggested by \cite{trainlonger:paper}, we also observe that training longer with large mini-batches helps
to reduce the training sub-optimality.

In addition to solving the system challenges of implementing large scale efficient synchronous DNN training,
another fundamental challenge for scaling synchronous DNN training is to make effective use
of an extremely large mini-batch in the optimization process.
\cite{goyal2017accurate} propose to increase learning rate linearly for larger mini-batch sizes.
\cite{lars:paper} propose to use layer specific gradient normalization to scale to larger mini-batch sizes.
\cite{progressivebatch:paper} propose to progressively increase the mini-batch size during the training process to
to control gradient noise levels in the early parts of the optimization process.
\cite{fastai} propose to use progressively larger images during the training process.
\cite{mixedprecision} propose to use mixed-precision during training to scale to large mini-batches.

In this work, we devise a novel pre-conditioned NLCG DNN training
algorithm with second order information that uses the large mini-batch sizes more effectively than
SGD based training algorithms.

\section{Nonlinear conjugate gradient method for DNN training} \label{sec:nlcg}

\subsection{DNN Training Optimization Background}

The objective of stochastic DNN training is to minimize the mini-batch loss
$L(\boldsymbol{w}) = \frac{1}{|X|} \displaystyle \sum_{\boldsymbol{x} \in X} l(\boldsymbol{x},\boldsymbol{w})$,
where $\boldsymbol{w}$ represents the model \emph{weights},
$X$ is a mini-batch, and $l(\boldsymbol{x},\boldsymbol{w})$ is the loss for a sample $\boldsymbol{x} \in X$.
Each iteration of stochastic gradient descent proceeds as follows: First, a mini-batch (a subset of the training data)
is sampled and used to compute a gradient estimate $\boldsymbol{g_t} = \nabla L(\boldsymbol{w_t})$,
where $t$ is the current training step.
The weights $\boldsymbol{w_t}$ are then updated with the
rule $\boldsymbol{w_{t+1}} = \boldsymbol{w_{t}} - \alpha \boldsymbol{g_t}$, where $\alpha > 0$ is a
hyper-parameter termed the \emph{learning rate}.
When the mini-batch is small, the variance of the gradient estimate is high. For the optimization to
succeed $\alpha$ should be small.

Popular variants of SGD generalize the above rule to
$\boldsymbol{w_{t+1}} = \boldsymbol{w_{t}} - \alpha \boldsymbol{d_t}$
where,
$\boldsymbol{d_t} = func(\boldsymbol{g_t}, \boldsymbol{g_{t-1}}, \cdots )$ which is a function of previous gradient estimates.
These methods include SGD with momentum (Momentum), RMSProp, Adam \cite{adam}, ADAGrad and others.
Since they only use gradient information (first derivative) they are referred to as first-order methods.

Newton's method achieves faster convergence than first order methods by leveraging information about the Hessian
of the target function. It models the function as a
quadratic by using a second order Taylor expansion at the current iterate. It then minimizes the approximation to choose
the next iterate. This results in the update vector $\boldsymbol{h} = \boldsymbol{H^{-1}g}$ with
($\boldsymbol{H} = \nabla^{2}L(\boldsymbol{w}),\ \boldsymbol{g} = \nabla L(\boldsymbol{w})$ ). Observe that the
computation of the update vector requires the inverse of the Hessian, or the solution of
the system $\boldsymbol{Hh} = \boldsymbol{g}$.

A model with $n$ weights results in a Hessian with $O(n^{2})$ entries. For a DNN with millions of weights,
computing, storing and solving $\boldsymbol{Hh} = \boldsymbol{g}$ is prohibitively expensive.
To this end, Hessian free optimization \cite{hfcg} uses Krylov-subspace methods to solve the linear system without ever
forming $\boldsymbol{H}$. Variants of these methods use linear conjugate gradient (CG), conjugate residuals (CR) and others.
Importantly, these methods only require forming products of the form $\boldsymbol{z} = \boldsymbol{Hy}$,
which can be achieved with the Pearlmutter trick \cite{pearlmutter}.

Second order methods have been previously used for DNN training \cite{deepoptimization}, \cite{lbfgs}
\cite{hfcg} \cite{dhir_hfcg}. However achieving good convergence results on
large scale DNN training problems like ImageNet classification has been
difficult.  Recently natural gradient based methods like K-FAC \cite{kfac} and
methods using Neumann power series \cite{neumann} have shown
promise in successfully training large scale problems like ImageNet
classification.

\subsection{Nonlinear Conjugate Gradient Method}
\label{sec:nlcg_algo}
Linear CG method is not only useful for solving linear systems, it is an optimization algorithm in its own right.
For a quadratic problem and a given accuracy threshold, CG will typically converge much faster than
gradient descent.

NLCG method generalizes the linear CG method to
nonlinear optimization problems and can also work successfully for non-convex optimization problems
\cite{boyd:opt}. However, it has not been explored successfully for large scale DNN training tasks
(e.g. ImageNet classification).
At each step $t$, NLCG chooses a direction, $\boldsymbol{d_t}$, that is conjugate ($\boldsymbol{H}$-orthogonal)
to all the previous directions (i.e. $\boldsymbol{d_t^T} \boldsymbol{H} \boldsymbol{d_{j}} = 0, \forall j < t$).
In contrast to gradient descent method, using conjugate gradients helps to avoid exploring
the same directions multiple times. It is possible to further improve convergence of the NLCG method by using
second order information through the pre-conditioner. We hypothesize that with large mini-batches,
the variance of the gradients is reduced, which makes NLCG method effective for DNN training.

We describe the stochastic preconditioned NLCG algorithm for DNN training in Algorithm \ref{alg:nlcg}.
The overall structure is very similar to the classic preconditioned NLCG algorithm \cite{Nocedal:opt}.
Key differences are the introduction of an efficient quasi newton pre-conditioner and online stochastic line search
to automatically determine step size.
At each iteration, we compute the gradient of the mini-batch using back-propagation.
As described in Algorithm \ref{alg:bfgs_precond} and Section \ref{sec:precond}, we compute a diagonal pre-conditioner
which estimates the curvature of the optimization problem at the current step. Using the diagonal pre-conditioner lets
us introduce second order information into the optimization process and reduces the condition number of the
system matrix to speed up convergence.

After pre-conditioning the gradient, we compute the conjugate direction using the Polak-Ribiere (PR) \cite{polakribiere} or
Fletcher-Reaves (FR) \cite{fletcherreaves} update formula to compute the $\beta$ term in Algorithm \ref{alg:nlcg}.
Like momentum term in the Momentum optimizer, $\beta$ determines
the amount of previous direction to keep in the new update. The conjugate gradient update formulas, PR and FR, are
designed to approximately keep each direction Hessian conjugate to previous directions.
Conjugate gradient methods help in efficiently traversing optimization landscapes with narrow ill-conditioned valleys
where gradient descent method can slow down.

\begin{algorithm}[tb]
   \caption{Nonlinear Conjugate Gradient Optimizer for DNN Training}
   \label{alg:nlcg}
\begin{algorithmic}
   \STATE $t = 0$
   \STATE $k = 0$
   \STATE $\boldsymbol{r} = - \nabla L(\boldsymbol{w})$  ~~~~~~~~// Compute Mini-Batch Gradient
   \STATE $\boldsymbol{M}^{-1}$ = Calculate Preconditioner $(\boldsymbol{w}, \nabla L(\boldsymbol{w}), t)$
   \STATE $\boldsymbol{s} = \boldsymbol{M}^{-1}\boldsymbol{r}$
   \STATE $\boldsymbol{d} = \boldsymbol{s}$
   \STATE $\delta_{new} = \boldsymbol{r}^{T}\boldsymbol{d}$
   \REPEAT
   \STATE $t = t + 1$
   \STATE $\alpha_t=\alpha^g_t*\alpha^s_t$ ~~~~~~// See Section \ref{sec:nlcg_ls} for $\alpha^g_t$, $\alpha^s_t$
   \STATE $\boldsymbol{w} = \boldsymbol{w} + \alpha_t \boldsymbol{d}$
   \STATE $\boldsymbol{r} = -\nabla L(\boldsymbol{w}) $  ~~~~~~// Compute Mini-Batch Gradient
   \STATE $\delta_{old} = \delta_{new}$
   \STATE $\delta_{mid} = \boldsymbol{r}^{T}\boldsymbol{s}$
   \STATE $\boldsymbol{M}^{-1}$ = Calculate Preconditioner$(\boldsymbol{w}, \nabla L(\boldsymbol{w}), t)$
   \STATE $\boldsymbol{s} = \boldsymbol{M}^{-1}\boldsymbol{r}$
   \STATE $\delta_{new} = \boldsymbol{r}^{T}\boldsymbol{s}$
   \IF {$update = PolakRibiere$}
   \STATE $\beta = \frac{\delta_{new} - \delta_{mid}}{\delta_{old}}$
   \ELSIF {$update = FletcherReaves$}
   \STATE $\beta = \frac{\delta_{new}}{\delta_{old}}$
   \ENDIF
   \IF {$\beta < 0$}
   \STATE $\beta = 0$
   \ELSIF {$\beta > 1$}
   \STATE $\beta = 1$
   \ENDIF
   \STATE $\boldsymbol{d} = \boldsymbol{s} + \beta \boldsymbol{d}$
   \UNTIL{$t < t_{max}$}
\end{algorithmic}
\end{algorithm}

\subsubsection{Quasi Newton Pre-conditioner}
\label{sec:precond}
Typically, the pre-conditioner in the NLCG method is supposed to approximate the Hessian inverse ($\boldsymbol{H}^{-1}$) of the optimization
problem. In addition, for convergence guarantees, the pre-conditioner is supposed to be constant during the optimization process.
For DNN training problems with several million optimization variables, computing the full Hessian inverse approximation
matrix would be prohibitive in both runtime and memory. It is non-trivial to compute a useful
static pre-conditioner for DNN training optimization. In our implementation, we use a dynamic pre-conditioner
that is updated using the quasi Newton BFGS update \cite{Nocedal:opt} at every iteration. The BFGS method approximates the
Hessian using the past difference of gradients and difference of weights to satisfy the secant equation
$\boldsymbol{H_{t+1}} (\boldsymbol{w_{t+1}} - \boldsymbol{w_{t}}) = \nabla L(\boldsymbol{w_{t+1}}) - \nabla L(\boldsymbol{w_{t}}) $
\cite{Nocedal:opt}. Our pre-conditioner is limited to only
the diagonal of the Hessian inverse which effectively scales the gradient of each variable individually.
By only computing the diagonal, we don't need to represent the BFGS approximate Hessian
matrix in memory. The inverse of the diagonal Hessian is also easy to compute. All the numerical operations involved
are vector additions, subtractions, multiplications or dot products, which keeps the runtime overhead of the diagonal
BFGS pre-conditioner low.

\begin{algorithm}[tb]
   \caption{Calculate Preconditioner: BFGS Diagonal Preconditioner $\approx H^{-1}(\boldsymbol{w})$}
   \label{alg:bfgs_precond}
\begin{algorithmic}
   \STATE {\bfseries Input:} variable $\boldsymbol{w}$, gradient $\nabla L(\boldsymbol{w})$, step $t$
   \IF{$t == 0$}
   \STATE $\boldsymbol{H}_{diag} = Identity$
   \ELSE
   \STATE $\boldsymbol{y} = \nabla L(\boldsymbol{w}) - \nabla L_{old}(\boldsymbol{w_{old}})$
   \STATE $\boldsymbol{s} = \boldsymbol{w} - \boldsymbol{w_{old}}$
   \STATE $\boldsymbol{H}_{diag} = \boldsymbol{H}_{diag} + \frac{Diag(\boldsymbol{yy^{T}})}{\boldsymbol{y^{T}s}} - \frac{\boldsymbol{H}_{diag} ~ Diag(\boldsymbol{ss^T}) ~ \boldsymbol{H}_{diag}^{\boldsymbol{T}}}{\boldsymbol{s^T} ~ \boldsymbol{H}_{diag} ~ \boldsymbol{s}}$
   \ENDIF
   \STATE $\boldsymbol{w_{old}} = \boldsymbol{w}$
   \STATE $\nabla L_{old}(\boldsymbol{w_{old}}) = \nabla L(\boldsymbol{w})$
   \STATE return $\boldsymbol{H}_{diag}^{-1}$
\end{algorithmic}
\end{algorithm}

\subsubsection{Online Stochastic Line Search}
\label{sec:nlcg_ls}
\begin{figure}
  \includegraphics[width=\linewidth]{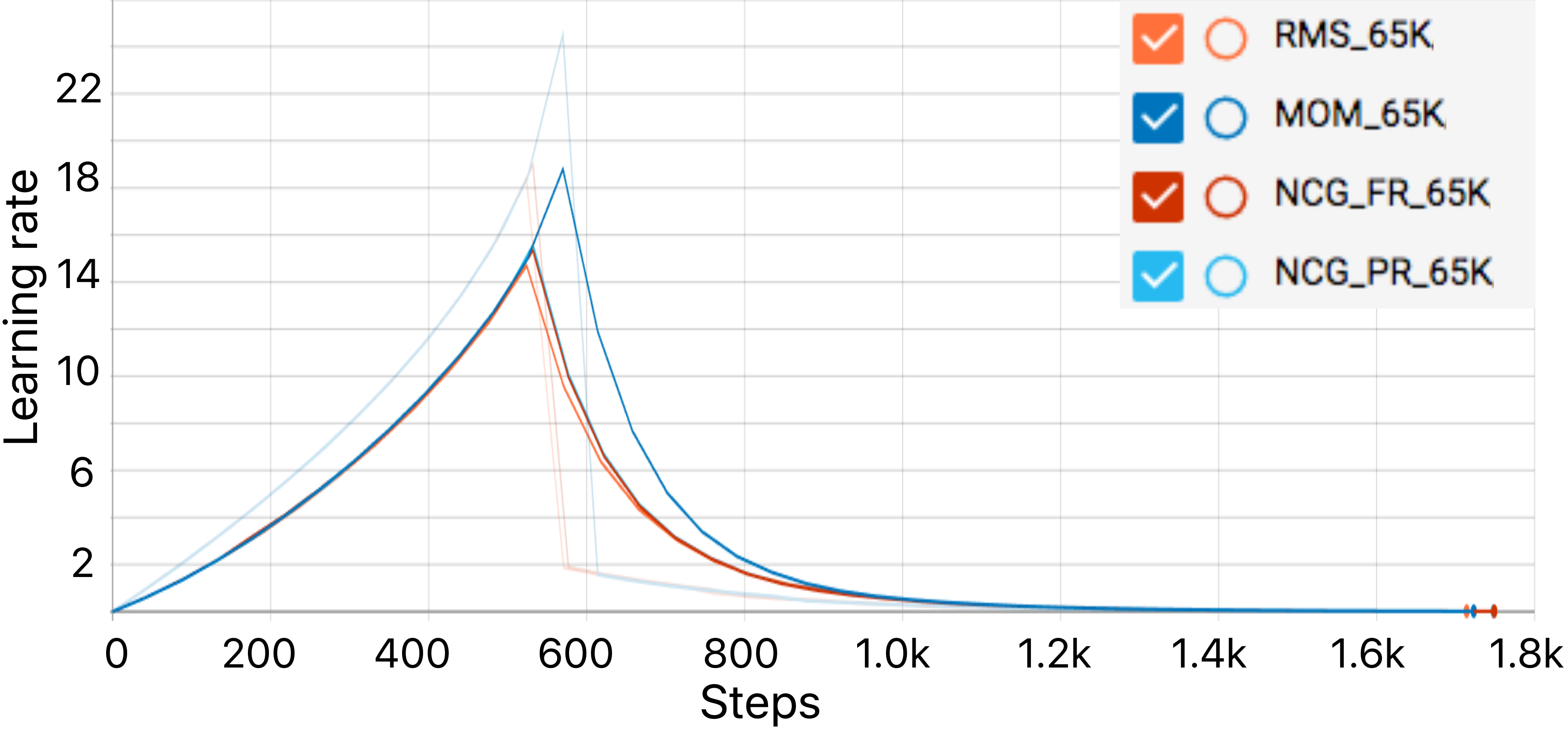}
  \caption{Learning rate schedule for training Resnet-50 model on ImageNet dataset with
  batch size 65536 using Momentum, RMSProp, NLCG\_FR (FletcherReaves), NLCG\_PR (PolakRibiere) optimizers.
  Learning rate schedules are identical for
  all optimizers. Differences in the chart are because of sampling / viewing artifacts of Tensorflow. }
  \label{fig:nlcg_learning_rate}
\end{figure}

Once the direction of the step is computed, we need to compute the step length. In classical NLCG, a line search procedure
\cite{Nocedal:opt} is employed to find the minimum of the function along the search direction. However, in a stochastic
setting, it is difficult to get a stable value of the function/gradient because of high variance introduced
by mini-batches. Even if the function value is stable, computing the step
length using traditional line search methods like secant line search \cite{boyd:opt} or
Armijo line search \cite{boyd:opt} can be very
expensive because several function and/or gradient calls are required. Instead,
we compute the global step length $\alpha^g_t$ by following a traditional learning rate schedule used in DNN training
(see Figure \ref{fig:nlcg_learning_rate}). We also compute a learning rate scale $\alpha^s_t$ which is multiplied
with global learning rate $\alpha^g_t$ to get the final learning rate $\alpha_t$. The learning rate scale $\alpha^s_t$ is
initialized to 1.0. 
We monitor the loss function value at each iteration update. If the loss increases by more than a certain
percentage (e.g. 2\%), we reduce the $\alpha^s_i$ by a decrease factor (e.g. 2.5\%). Conversely, if the loss  reduces or
stays flat (e.g. <1\% increase), we increase $\alpha^s_t$ by an increase (e.g. 2.5\%), up to the maximum of 1.0. 
Thresholds used for $\alpha^s_t$ computation are hyper-parameters that depend on the variance of the loss function and may be 
tuned for better performance. The online stochastic line search strategy is inspired by the back-tracking line search strategy used in convex
optimization. \footnote{\label{opensource} Reference open-source NLCG tensorflow implementation available at \textbf{https://github.com/apple/ml-ncg}}

\section{Results}\label{sec:results}

In this section, we detail the results of using the NLCG method to train the Resnet-50 model on the ImageNet image classification
task and to train the Resnet-32 model on the CIFAR-100 image classification task.

ImageNet image classification task is a large scale learning task and deep learning models have been successful
in obtaining state of the art accuracy on this classification task.
We conduct our experiments in the large mini-batch scenario with 299x299 image crops.
We use parameter server based distributed DNN training.
Recent work on synchronous distributed training have shown scaling results upto mini-batch size of 65536 using 1024 GPUs
to obtain very fast training times.
We restrict our experiments to 64 V100 GPUs, primarily due to availability constraints in our cluster.
One Nvidia V100 GPU can efficiently process 64 299x299 image crops from the ImageNet training dataset.
Therefore the maximum native mini-batch size we can experiment with is 4096.
For batch sizes greater than 4096, we use virtual batching to simulate the effect of larger mini-batches.
In virtual batching, each worker uses multiple
mini-batches to compute gradients and averages them before communicating the update to the parameter server.
The parameter server then computes an average of averaged gradients before applying the update to the shared
weight parameters. This effectively increases the mini-batch size for a fixed number of workers.
Virtual batching increases the system computation to communication ratio and improves distributed training
system throughput.



Each step of preconditioned NLCG optimizer does more work than traditional SGD, SGD with Momentum (Momentum) or RMSProp
with Momentum (RMSProp) optimizers. Hence, the system throughput is expected to be lower than traditional SGD based training
methods.
In our experiments, a single NLCG training step is about 15\% to 30\% slower than natively
implemented RMSProp/Momentum.
Note that, due to virtual batching, this overhead is not visible in Figure \ref{fig:nlcg_learning_curve_walltime}.
This is because of higher computation cost of computing average gradient of the
virtual mini-batch compared to the additional computational overhead of the NLCG optimizer.

\begin{figure}
  \includegraphics[width=\linewidth]{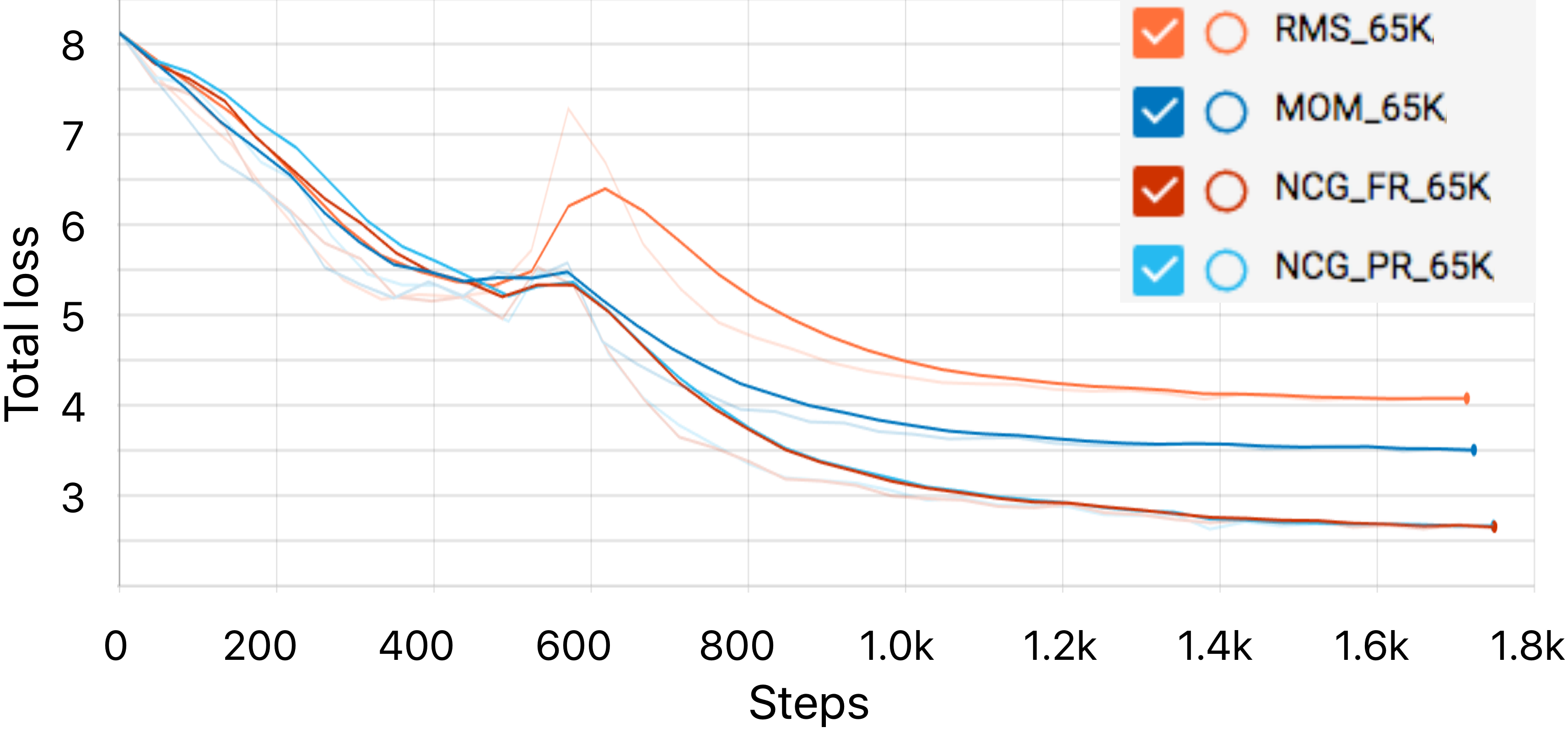}
  \caption{Training loss function convergence curve per step for training Resnet-50 with a batch size 65536
  on the ImageNet dataset using Momentum, RMSProp, NLCG\_FR and NLCG\_PR optimizers. NLCG optimizers converge to
  solutions with better training loss.}
  \label{fig:nlcg_learning_curve}
\end{figure}

\subsection{Convergence studies}

The initial learning rate is set to 0.001.
We employ standard DNN training with learning rate increase with warmup \cite{goyal2017accurate} for 5 epochs
(max learning rate = $0.1*\frac{BatchSize}{256}$ ).
For mini-batch sizes less than 8192,
we reduce the learning rate to 0.001 using exponential rate decay every 2 epochs. For mini-batch sizes greater than or equal to 8192, we
reduce the learning rate to 0.01 using exponential rate decay every 2 epochs.
For mini-batch sizes greater than 8192, the warm up epochs is increased to 15.
For mini-batch sizes greater than 32768, the warm up epochs is increased to 30.
To be consistent in the experiments, we keep the same learning rate schedule for all optimizers, which is tuned for Momentum optimizer.
With this setting, we are able to achieve the state of the art top-1 accuracy for the baseline
SGD based optimizers.
For illustration, we detail the learning rate schedule curve for
NLCG, RMSProp and Momentum optimizers with a batch size of 65536 in Figure \ref{fig:nlcg_learning_rate}.
We use a momentum value of 0.9 for both RMSProp and Momentum optimizers.
We use standard image augmentation pipeline with 299x299 image crops.

With these settings our baseline results for Momentum and RMSProp optimizers are similar to the work in
\cite{goyal2017accurate} which is closest to our setting of traditional DNN training.
We do not use techniques customized for large batch ImageNet training from recent works like LARS\cite{lars:paper},
progressive batch size increases\cite{progressivebatch:paper} and progressive image size increases\cite{fastai}).
We study the performance of various optimizers in the standard DNN training setting.

\begin{figure}
  \includegraphics[width=\linewidth]{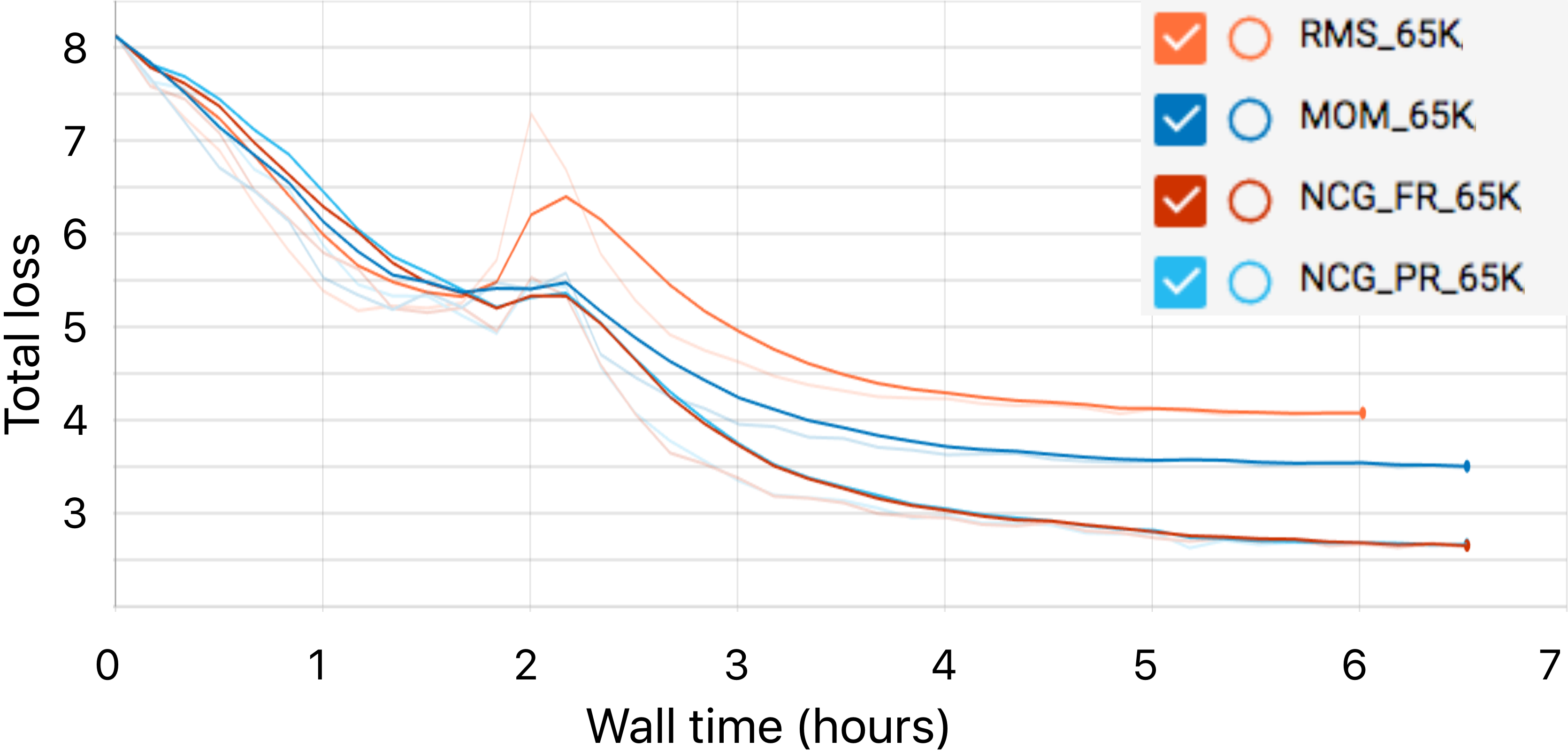}
  \caption{Training loss function convergence curve per wall-clock time for training Resnet-50 model with batch size 65536
  on ImageNet dataset with Momentum, RMSProp, NLCG\_FR and NLCG\_PR optimizers. NLCG optimizers converge to solutions with
  better training loss in similar amount of walltime.
  We train with 64 V100 GPUs and use virtual batching to simulate the high batch size of 65536.
  Virtual batching reduces the runtime overhead associated with NLCG optimizer.
  Wall-clock times can be noisy because of variations in (network communication load, machine placement, etc) in the
  cluster environment.
  }
  \label{fig:nlcg_learning_curve_walltime}
\end{figure}

In Figures \ref{fig:nlcg_learning_curve} and \ref{fig:nlcg_learning_curve_walltime}, we show the training loss function convergence curves for
NLCG\_FR, NLCG\_PR, RMSProp and Momentum optimizers at an extremely large mini-batch size of 65536. We observe that at these extremely large
mini-batch sizes, preconditioned NLCG optimizers have better loss convergence per step. The final loss value achieved
by NLCG methods is also better than Momentum or RMSProp.

\begin{figure}
  \includegraphics[width=\linewidth]{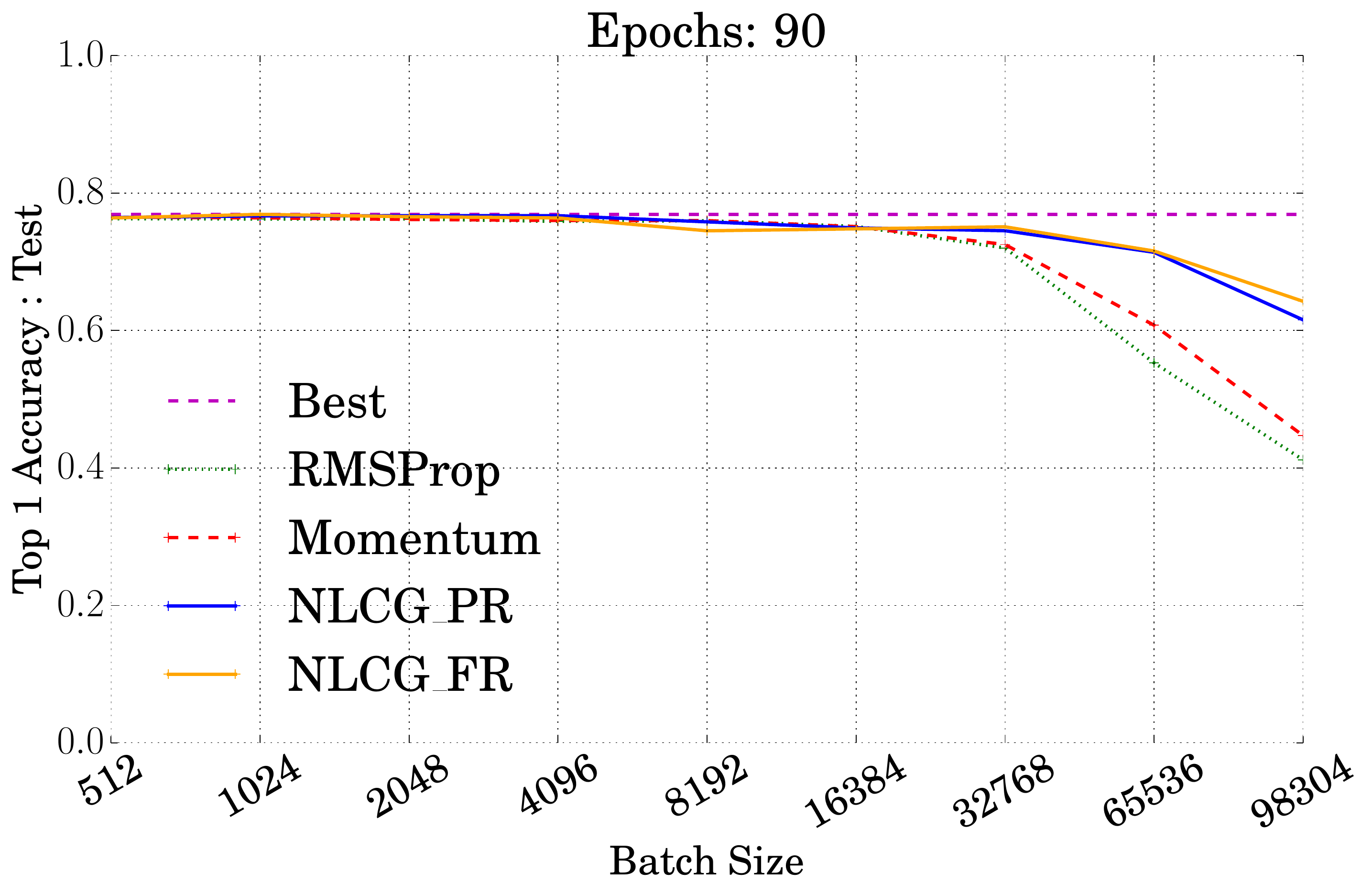}
  \caption{Top-1 accuracy of Resnet-50 model trained for 90 epochs on ImageNet data using NLCG\_PR(PolakRibiere), NLCG\_FR(FletcherReaves),
  Momentum, RMSProp optimizers.
  We apply standard training procedure and show results for batch sizes from 512 to 98304. Best line is for comparison and
  represents the best top-1 accuracy (76.9\%) achieved by any optimizer (NLCG\_FR) at any batch size (1024) in the study.
  NLCG optimizers dominate the Momentum and RMSProp for mini-batch sizes greater than 16384.}
  \label{fig:accuracy_vs_batchsize}
\end{figure}

In Figure \ref{fig:accuracy_vs_batchsize}, we show the top-1 test accuracy of Resnet-50 models trained using various
optimizers. We train the Resnet-50 model on ImageNet training data for 90 epochs and vary the mini-batch size
from 512 to 98304. We only concentrate on large and extremely large mini-batch sizes for this study.
For batch sizes 16384 and lower, all the optimizers are able to get a top-1 accuracy close to or greater than 75\%, with the best
accuracy being 76.9\% obtained by NLCG\_FR optimizer at batch size of 1024.
For comparison, we show the best baseline top-1 accuracy (76.9\%) as a separate line.
In this study, we observe that as the batch size increases >16384, the four optimizers under study start degrading the top-1 test
accuracy. Amongst the 4 optimizers, preconditioned NLCG optimizers drop the accuracy the least. This is in line with
the observation in Figure \ref{fig:nlcg_learning_curve} that with NLCG optimizers, we optimize the training loss function
better to achieve a lower loss function value.
At batch size of 65536, the difference in accuracy between NLCG\_FR and Momentum is about 10.3\%.
At batch size of 98304, the difference in accuracy between NLCG\_FR and Momentum is about 19.5\%.
For Resnet-50 training on ImageNet data, with fixed number of epochs, NLCG optimizers provide a lower training loss
for large batch sizes.

\begin{figure}
  \includegraphics[width=\linewidth]{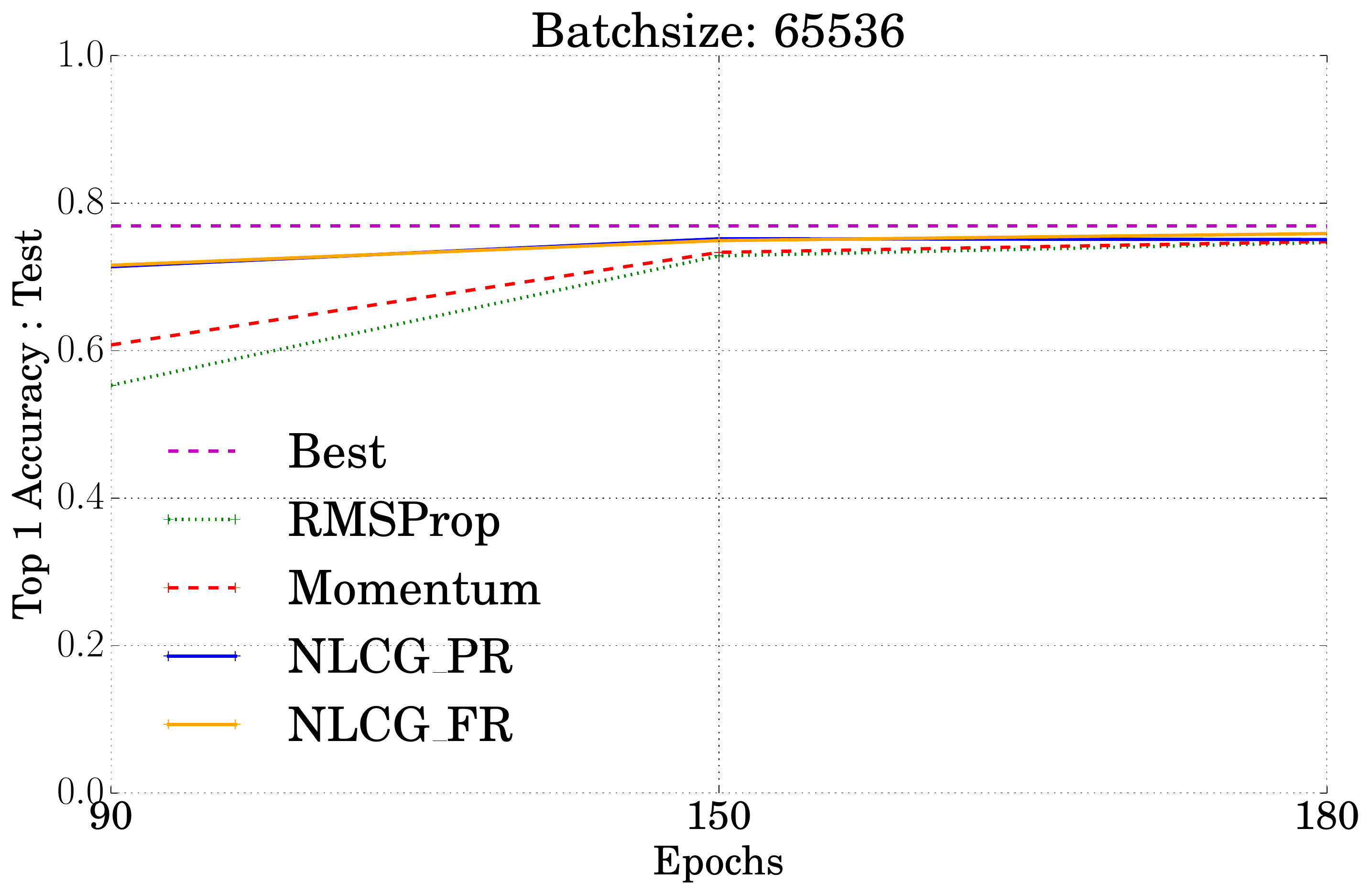}
  \caption{Top-1 accuracy of Resnet-50 model trained with a batch size of 65536 on ImageNet data using NLCG\_PR(PolakRibiere),
  NLCG\_FR(FletcherReaves), Momentum, RMSProp optimizers. We apply standard training procedure and show results for 90,
  150 and 180 epochs. Best line is for comparison and
  represents the best top-1 accuracy (76.9\%) achieved by any optimizer (NLCG\_FR) at any batch size (1024) in the study.
  NLCG optimizers dominate Momentum and RMSProp for this mini-batch size as number of training epochs increase.}
  \label{fig:accuracy_vs_epoch_64k}
\end{figure}

\begin{figure}
  \includegraphics[width=\linewidth]{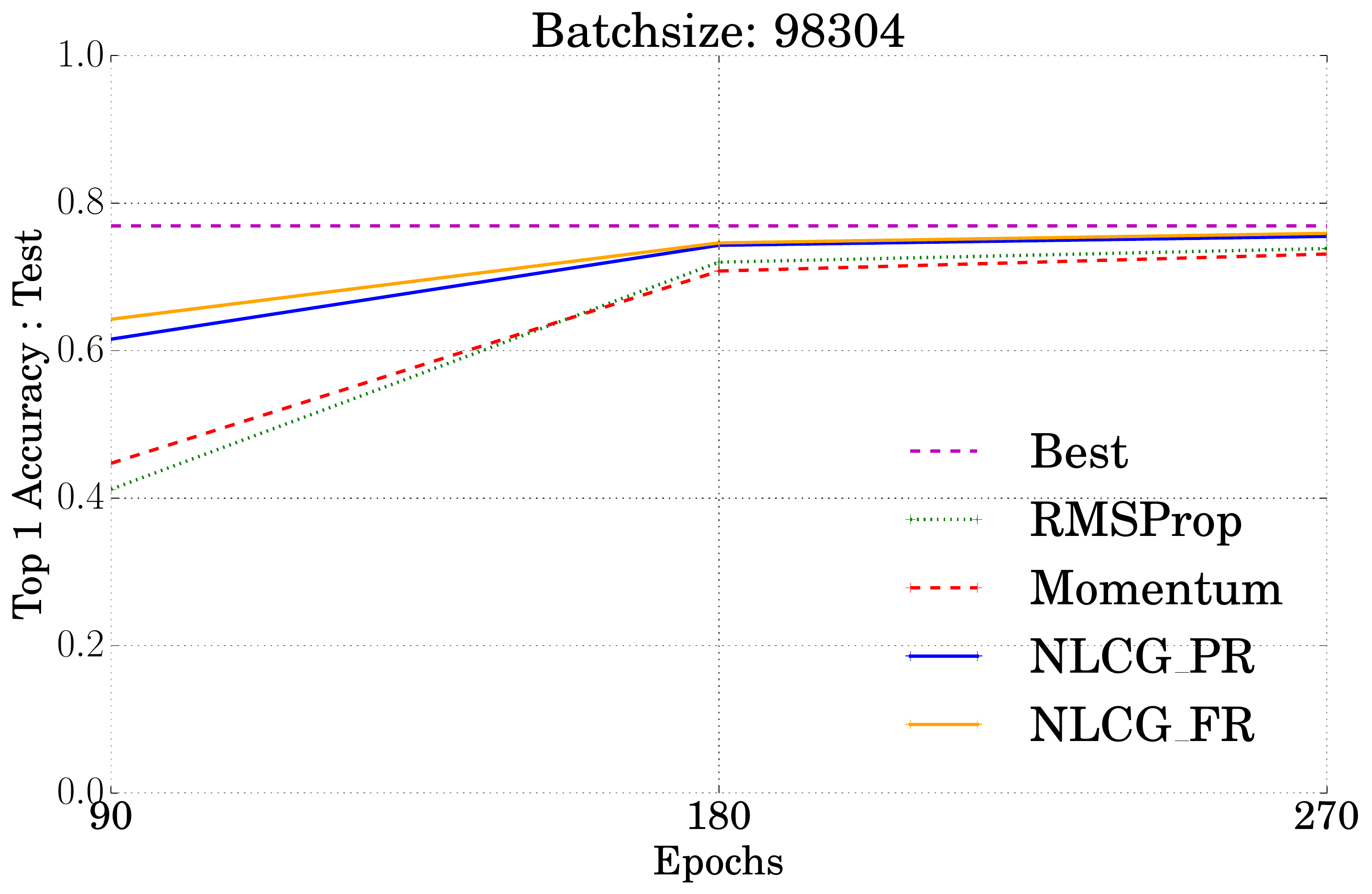}
  \caption{Top-1 accuracy of Resnet-50 model trained with a batch size of 98304 on ImageNet data using NLCG\_PR(PolakRibiere), NLCG\_FR(FletcherReaves),
  RMSProp optimizers. We apply standard training procedure and show results for 90, 180 and 270 epochs. Best line is for comparison and
  represents the best top-1 accuracy (76.9\%) achieved by any optimizer (NLCG\_FR) at any batch size (1024) in the study.
  NLCG optimizers dominate Momentum and RMSProp for this mini-batch size as number of training epochs increase.}
  \label{fig:accuracy_vs_epoch_98k}
\end{figure}

As seen in Figure \ref{fig:accuracy_vs_batchsize}, with large mini-batches of 65536 and 98304, we are not able to get close to
the baseline top-1 accuracy when training for 90 epochs. We want to understand if this is a result of training difficulty
or a generalization issue \cite{largebatch:paper} because of large batch sizes.
In Figure \ref{fig:accuracy_vs_epoch_64k}, we fix the batch size to 65536 and run the training longer.
We train for 90, 150 and 180 epochs to see if we can close the test accuracy gap by training longer.
For each experiment, we change the learning rate schedule so that the
final learning rate value at the end of the specified number of epochs is 0.01.
We observe that at 180 epochs, all 4 optimizers are able to get close to 75\% top-1 accuracy with NLCG optimizers dominating
amongst the 4 optimizers.
In Figure \ref{fig:accuracy_vs_epoch_98k}, we repeat the same study at an even larger batch size of 98304 and
train for 90, 180 and 270 epochs. We observe that at this batch
size only NLCG optimizers are able to reach the top-1 accuracy close to 75\% by training longer.
Our conclusion is similar to \cite{trainlonger:paper} that for extremely large mini-batches, training longer helps in reducing training
sub-optimality and improves test accuracy. As the batch sizes grow, the preconditioned NLCG optimizers starts
to dominate the Momentum and the RMSProp optimizers in terms of loss/accuracy convergence per step.
Note that the ImageNet convergence/accuracy graphs are generated using Resnet-50 V2 architecture.
We repeated the study also on training Resnet-50 V1 architecture and reached the same conclusion.

\begin{figure}
  \includegraphics[width=\linewidth]{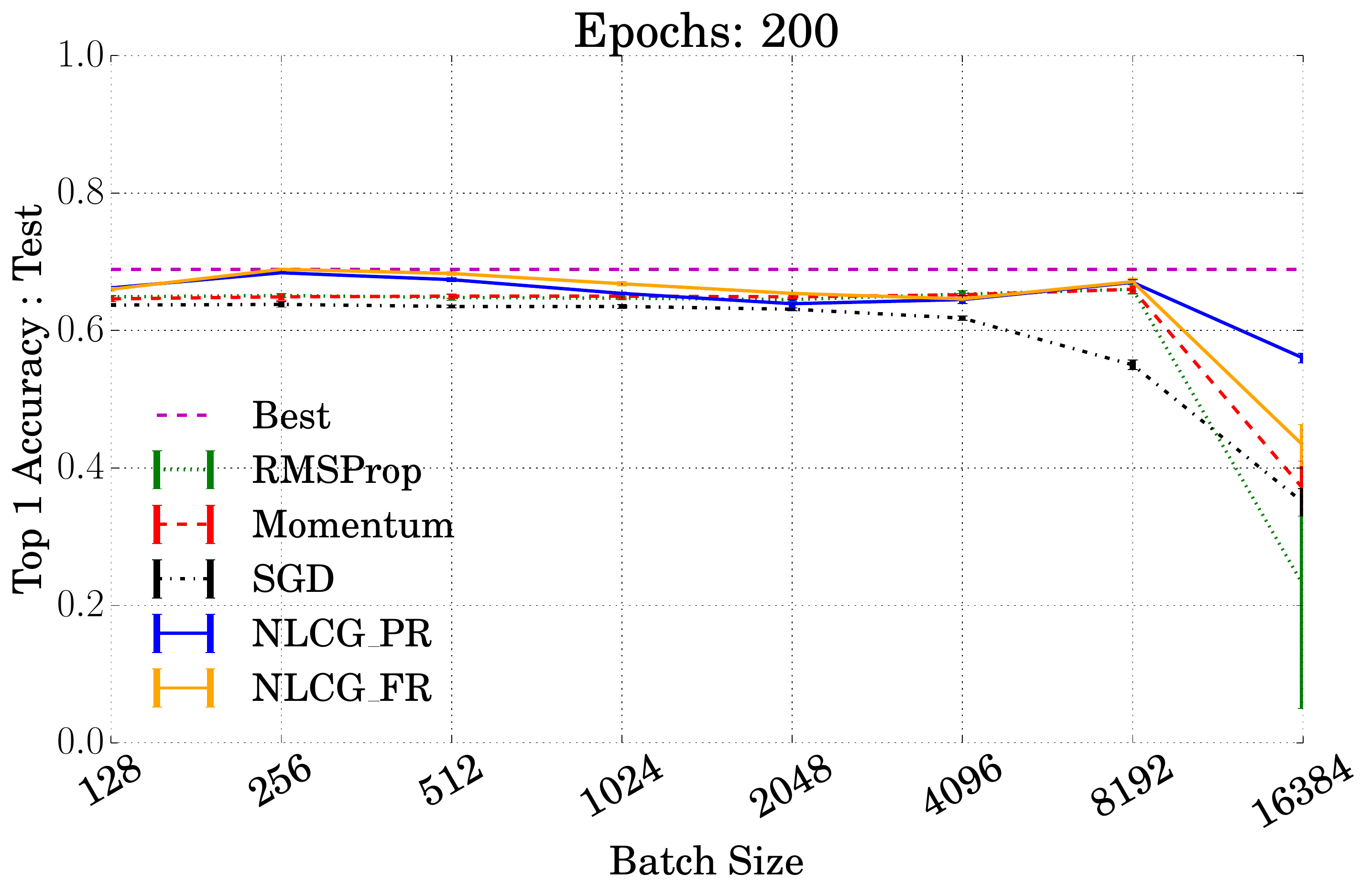}
  \caption{Top-1 accuracy of Resnet-32 model trained for 200 epochs on CIFAR-100 data using NLCG\_FR(FletcherReaves),
   NLCG\_PR(PolakRibiere),
   Momentum, RMSProp and SGD optimizers. We apply standard training procedure and show results for batchsizes
   from 128 to 16384. We show mean and standard deviation error bars for 5 experiments per point. Best line is for comparison and
   represents the best top-1 accuracy (68.9\%) achieved by any optimizer (NLCG\_FR) and any batch size (256) in the study.
   NLCG optimizers dominate SGD, Momentum and RMSProp for mini-batch sizes greater than 8192.}
  \label{fig:cifar100_accuracy_vs_batchsize}
\end{figure}

\begin{figure}
  \includegraphics[width=\linewidth]{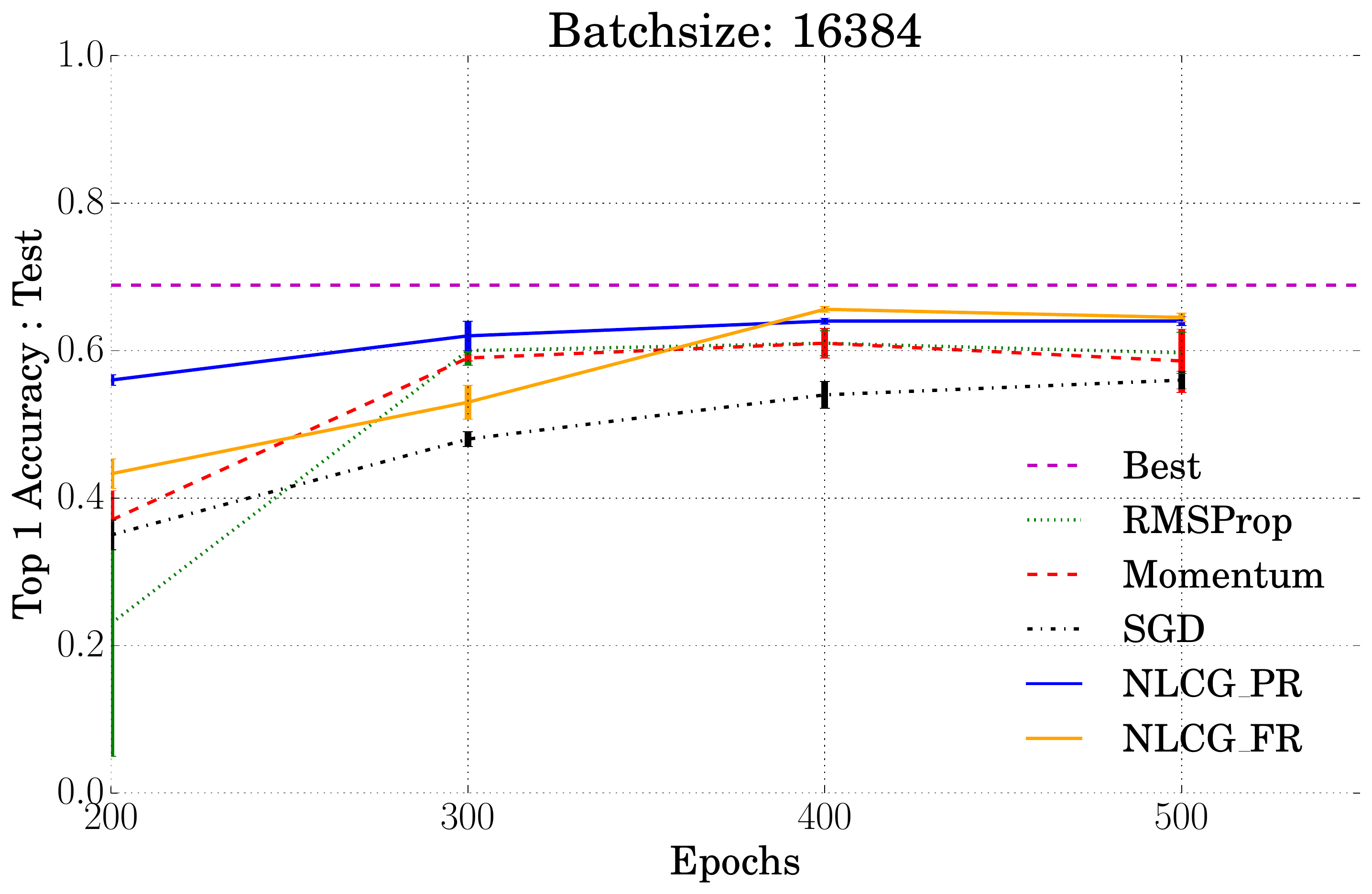}
  \caption{Top-1 accuracy of Resnet-32 model trained with a mini-batch size of 16384 on CIFAR-100 data using
  NLCG\_FR(FletcherReaves), NLCG\_PR(PolakRibiere), Momentum, RMSProp and SGD optimizers.
  We apply standard training procedure and show results for 200, 300, 400 and 500 epochs.
  We show mean and standard deviation error bars for 5 experiments per point. Best line is for comparison and
  represents the best top-1 accuracy (68.9\%) achieved by any optimizer (NLCG\_FR) and any batch size (256) in the study.
  NLCG optimizers dominate SGD, Momentum and RMSProp for this mini-batch sizes as we increase the training epochs.}
  \label{fig:cifar100_accuracy_vs_epoch_16k}
\end{figure}

To study the performance of the NLCG optimizer on a different dataset, we repeat the experiments on training the Resnet-32 model on
the CIFAR-100 data. The results are detailed in Figures \ref{fig:cifar100_accuracy_vs_batchsize}
and \ref{fig:cifar100_accuracy_vs_epoch_16k}. We run the training for 200 epochs and report the top-1 accuracy
for NLCG\_FR, NLCG\_PR, Momentum, RMSProp and SGD optimizers. We run 5 experiments for each data point and show the mean and the standard
deviation error bars. For comparison, we also show the best top-1 accuracy (68.9\%) achieved by any optimizer (NLCG\_FR) at any mini-batch
size (256). For the NLCG optimizers, we had to turn off the line search for mini-batch sizes less than 2048, because the mini-batch
loss function value was too noisy to do meaningful loss function monitoring.
We observe that as the mini-batch sizes increase to 16384, all optimizers start degrading the top-1 accuracy. NLCG\_PR and
NLCG\_FR optimizers are able to maintain the best top-1 accuracy when training for a fixed number of epochs (200).

In Figure \ref{fig:cifar100_accuracy_vs_epoch_16k}, we fix the mini-batch size to 16384 and progressively increase the number of training
epochs from 200 to 500. We observe that NLCG based optimizers are able to obtain the best accuracies. The standard deviations for
some points are large because some of the runs can diverge with such a large batch size. Over the several experiments,
NLCG optimizers seems to be more robust to such divergences.

\section{Conclusion and Future Work}\label{sec:future_work}
Training with large mini-batches is
essential to scaling up synchronous distributed DNN training.
We proposed a novel NLCG based optimizer that uses second order information to scale DNN training
with extremely large mini-batches.
The NLCG optimizer uses conjugate gradients and an efficient diagonal pre-conditioner to speed-up the
training convergence.
We demonstrated on the ImageNet and the CIFAR-100 datasets that for large mini-batches,
our method outperforms existing state-of-the-art optimizers by a large margin.

There is a runtime overhead in the NLCG optimizer because of more processing being done to compute
the pre-conditioned search direction.
For this work, we have implemented the preconditioned NLCG optimizer in TensorFlow using several TensorFlow ops.
We can further improve the runtime of NLCG optimizer by natively implementing the
optimizer as a fused native TensorFlow op with an optimized C++ / CUDA implementation.

We have mainly focused this study on the large mini-batch scenario. The NLCG method is stable in this scenario because the
gradients have low variance and are meaningful to compute valid pre-conditioners and conjugate search directions.
As the mini-batch becomes very small (eg. 128), the variance of gradients increase and the NLCG
method becomes unstable (preconditioning, conjugate search direction computation and line search). We intend to research
techniques to stabilize the NLCG method in the smaller mini-batch, high gradient variance scenario.

\section*{Acknowledgements}
We would like to thank Ashish Shrivastava, Dennis De Coste, Shreyas Saxena, Santiago Akle,
Russ Webb, Barry Theobald and Jerremy Holland for valuable comments for manuscript preparation.





\bibliography{bib/sysml_paper}
\bibliographystyle{bib/sysml2019}





\end{document}